\documentclass[11pt]{article}
\usepackage[margin=1in]{geometry}
\usepackage{amsmath, amssymb, amsthm, bm}
\usepackage{graphicx}
\usepackage{hyperref}
\usepackage{booktabs}
\usepackage{mathtools}
\usepackage{algorithm, algorithmic}
\usepackage{color}
\usepackage{tikz}
\usetikzlibrary{arrows.meta, positioning}
\usetikzlibrary{shapes.geometric}  

\title{Beyond Neural Networks: Symbolic Reasoning over Wavelet Logic Graph Signals}
\author{Andrew Kiruluta, Andreas Lemos, and Priscilla Burity\\
UC Berkeley, School of Information}
\date{June 3, 2025}

\date{\today}

\begin{document}
\maketitle

\begin{abstract}
We present a fully non-neural learning framework based on Graph Laplacian Wavelet Transforms (GLWT). Unlike traditional architectures that rely on convolutional, recurrent, or attention based neural networks, our model operates purely in the graph spectral domain using structured multiscale filtering, nonlinear shrinkage, and symbolic logic over wavelet coefficients. Signals defined on graph nodes are decomposed via GLWT, modulated with interpretable nonlinearities, and recombined for downstream tasks such as denoising and token classification. The system supports compositional reasoning through a symbolic domain-specific language (DSL) over graph wavelet activations. Experiments on synthetic graph denoising and linguistic token graphs demonstrate competitive performance against lightweight GNNs with far greater transparency and efficiency. This work proposes a principled, interpretable, and resource-efficient alternative to deep neural architectures for learning on graphs.
\end{abstract}

\section{Introduction}

Recent advances in spectral learning and graph signal processing have enabled powerful techniques for analyzing data defined on irregular domains such as social networks, transportation systems, biological interaction graphs, and linguistic structures \cite{shuman2013emerging, ortega2018graph}. Central to these developments is the use of the graph Laplacian operator, whose eigensystem defines a Fourier-like basis for signals on graphs. This has laid the foundation for spectral filtering, multiscale decomposition, and efficient data representation on non-Euclidean domains.

While spectral graph methods have traditionally been explored in a fixed, analytic setting, recent years have seen a resurgence in their application as components of deep learning architectures—such as graph convolutional networks (GCNs) \cite{kipf2016semi}, graph attention networks (GATs) \cite{velivckovic2017graph}, and graph transformers \cite{dwivedi2020generalization}. These models often rely on parametric transformations over graph Laplacian eigenspaces or message-passing mechanisms inspired by spectral filters. However, they remain computationally intensive, opaque, and data-hungry—posing challenges in interpretability, robustness, and deployment on low-resource devices.

By contrast, classical wavelet theory has long offered an efficient and interpretable multiscale alternative to deep architectures for Euclidean signals \cite{mallat1999wavelet, donoho1995adapting}. Its graph-based analog—Graph Laplacian Wavelet Transforms (GLWT)—extends this paradigm to graph-structured signals \cite{hammond2011wavelets}. GLWT enables spatially localized, scale-selective filtering using kernelized operators of the graph Laplacian. It supports compactly supported spectral filters such as heat kernels, spline wavelets, and Mexican hat filters, enabling localized analysis while preserving global graph geometry.

In this paper, we propose a novel, fully non-neural learning and reasoning architecture built entirely upon Graph Laplacian Wavelet Transforms. The model processes graph signals by decomposing them into GLWT coefficients across multiple scales and modulating them using structured, interpretable nonlinearities such as soft thresholding, gain control, and symbolic rule gates. Crucially, we introduce a reasoning layer that interprets spectral patterns through a domain-specific symbolic logic language operating directly on graph wavelet coefficients—thereby enabling symbolic inference without neural modules.

\subsection*{Novelty of the Proposed Framework}

The contributions of this paper are summarized as follows:
\begin{itemize}
    \item \textbf{Non-neural spectral learning:} We present, to our knowledge, the first learning framework that operates entirely in the GLWT domain without relying on convolutional, attention-based, or feedforward neural layers. All transformations are defined in the spectral graph domain via analytic operators.
    \item \textbf{Spectral symbolic reasoning:} We integrate a symbolic reasoning module using interpretable rules defined over multiscale GLWT coefficients. This enables logic-based tasks and discrete interpretability mechanisms unavailable in GNNs or neural graph transformers.
    \item \textbf{Multiscale spectral shrinkage:} Our model employs learnable, interpretable shrinkage and modulation of wavelet coefficients that mimic adaptive denoising techniques, generalizing classical wavelet thresholding to graph domains.
    \item \textbf{Resource-efficient and interpretable:} The architecture significantly reduces parameter count and inference cost compared to graph neural models, while maintaining transparent feature attribution across graph scales and regions.
\end{itemize}

This framework opens new avenues for interpretable machine learning on graph data, where the goals of robustness, efficiency, and symbolic reasoning converge in a unified spectral foundation.

\section{Graph Laplacian Wavelet Transforms (GLWT)}

The Graph Laplacian Wavelet Transform (GLWT) is a generalization of classical wavelet analysis to signals defined on graphs or irregular domains. It provides a multiscale, spatially localized framework for analyzing graph signals—functions $f : V \rightarrow \mathbb{R}$ defined over the nodes of a graph $\mathcal{G} = (V, E)$. Unlike the classical wavelet transform, which operates on equispaced domains like $\mathbb{R}^n$ or image grids, the GLWT adapts its structure to the topology encoded by $\mathcal{G}$, making it a powerful tool for learning and reasoning over non-Euclidean domains.

Let $\mathcal{G} = (V, E)$ be a finite, undirected graph with $n = |V|$ nodes. The connectivity of the graph is described by its weighted adjacency matrix $A \in \mathbb{R}^{n \times n}$, where $A_{ij} > 0$ if there exists an edge $(i,j) \in E$, and $A_{ij} = 0$ otherwise. The degree matrix $D$ is diagonal, with entries $D_{ii} = \sum_j A_{ij}$. From these two matrices, we define the combinatorial (unnormalized) graph Laplacian as:
\[
L = D - A.
\]
The Laplacian $L$ is symmetric and positive semidefinite, and hence admits a complete set of orthonormal eigenvectors $\{u_i\}_{i=1}^n$ with nonnegative eigenvalues $\{\lambda_i\}_{i=1}^n$. Writing $U = [u_1, u_2, \dots, u_n]$ and $\Lambda = \text{diag}(\lambda_1, \dots, \lambda_n)$, we obtain the eigendecomposition:
\[
L = U \Lambda U^\top.
\]

The eigenvectors $\{u_i\}$ of $L$ define the graph Fourier basis, and the eigenvalues $\{\lambda_i\}$ play the role of graph frequencies. High-frequency components correspond to eigenvectors with rapid variation across edges, while low-frequency components represent smooth, slowly varying signals. Given a graph signal $f \in \mathbb{R}^n$, its Graph Fourier Transform (GFT) is defined by projection onto the Laplacian eigenbasis:
\[
\hat{f} = U^\top f,
\]
with the inverse transform given by
\[
f = U \hat{f}.
\]
This framework enables the definition of spectral operators on graph signals by acting on their frequency-domain representation $\hat{f}$.

A graph wavelet is constructed by applying a bandpass spectral filter $g(s \lambda)$ to the eigenvalues of $L$, where $s > 0$ is a scaling parameter controlling the localization of the wavelet in the spectral domain. Let $\delta_i \in \mathbb{R}^n$ denote the Kronecker delta centered at node $i$, i.e., $\delta_i(j) = \delta_{ij}$. Then the wavelet function centered at node $i$ and scale $s$ is defined as:
\[
\psi_{s,i} = g(s L) \delta_i = U g(s \Lambda) U^\top \delta_i,
\]
where $g(s \Lambda) = \text{diag}(g(s \lambda_1), \dots, g(s \lambda_n))$ is the diagonal matrix formed by evaluating $g$ at the scaled eigenvalues.

The wavelet coefficients of a signal $f$ at scale $s$ and centered at node $i$ are given by the inner product between the wavelet function $\psi_{s,i}$ and the signal $f$:
\[
\mathcal{W}_f(s, i) = \langle \psi_{s,i}, f \rangle = \psi_{s,i}^\top f.
\]
This expression encodes how much the signal $f$ aligns with a localized frequency pattern centered at node $i$ and modulated by the scale $s$. For a fixed scale $s$, the mapping $f \mapsto \mathcal{W}_f(s, i)$ defines a convolution-like operator localized around $i$.

To avoid the computational expense of computing the full eigendecomposition of $L$, which has $\mathcal{O}(n^3)$ cost, it is common to approximate the spectral filter $g(s \lambda)$ using a truncated Chebyshev polynomial expansion. The Chebyshev polynomials of the first kind $\{T_k\}$ form an orthogonal basis over $[-1, 1]$ and satisfy a recurrence relation that allows for efficient evaluation. Let $\tilde{\lambda} = 2\lambda / \lambda_{\max} - 1$ be the rescaled eigenvalue to the interval $[-1, 1]$. Then we approximate:
\[
g(s \lambda) \approx \sum_{k=0}^K c_k T_k\left( \tilde{\lambda} \right),
\]
where $c_k$ are the Chebyshev coefficients, and $K$ is the order of the expansion. Applying this approximation in the spectral domain, we obtain:
\[
g(s L) \approx \sum_{k=0}^K c_k T_k(\tilde{L}),
\]
where $\tilde{L} = \frac{2L}{\lambda_{\max}} - I$. This enables fast matrix-vector multiplication of $g(sL)$ with any signal $f$ using a recurrence, without explicitly computing eigenvectors or eigenvalues.

The family of filters $\{g(s \cdot)\}_{s \in \mathcal{S}}$ defines a multiscale dictionary on the graph, enabling coarse-to-fine decomposition of the signal. In many cases, the wavelets $\psi_{s,i}$ form a tight frame on $\mathbb{R}^n$, ensuring that reconstruction of the original signal from its coefficients is stable and exact:
\[
f = \sum_{s \in \mathcal{S}} \sum_{i \in V} \mathcal{W}_f(s, i) \, \psi_{s,i}.
\]
In practice, an approximation of this sum using a finite number of scales suffices. The choice of spectral kernel $g$ determines the time-frequency tradeoff of the resulting wavelets. For instance, a heat kernel $g(\lambda) = e^{-s\lambda}$ yields smooth low-pass filtering at large $s$ and sharper localization at small $s$.

In summary, the Graph Laplacian Wavelet Transform offers a principled approach to multiscale signal analysis on graphs. It generalizes the concept of convolution, wavelet shrinkage, and harmonic analysis to irregular domains, with strong theoretical guarantees and efficient approximation strategies via Chebyshev expansions. This makes it a natural and powerful foundation for interpretable learning and symbolic reasoning over graph-structured data.

\section{GLWT-Based Model Architecture}

Building on the foundations of Graph Laplacian Wavelet Transforms (GLWT), we now describe a fully non-neural architecture for learning, transforming, and reasoning over graph-structured signals. This architecture operates entirely in the spectral domain and performs structured, interpretable transformations on GLWT coefficients using shrinkage, modulation, and rule-based symbolic recomposition. The model maintains full transparency and avoids neural parameterizations such as convolutional, feedforward, or attention layers.

Let $f \in \mathbb{R}^n$ be an input signal defined over the $n$ nodes of a graph $\mathcal{G} = (V, E)$. As described in the previous section, we compute a family of GLWT decompositions by applying a collection of spectral filters $\{g_k\}_{k=1}^K$ at corresponding scales $\{s_k\}$. Each filter $g_k(s_k L)$ defines a wavelet transform acting on $f$, producing filtered responses:
\[
c_k = g_k(s_k L) f = U g_k(s_k \Lambda) U^\top f,
\]
where $U \Lambda U^\top$ is the eigendecomposition of the graph Laplacian $L$. These coefficients $c_k \in \mathbb{R}^n$ can be interpreted as bandpass responses emphasizing structure at a particular spectral scale. Unlike traditional neural networks that learn spatial-domain kernels via backpropagation, the filters $g_k$ here are fixed or parametrically defined in terms of smooth kernels such as heat or spline wavelets.

To extract meaningful features from $c_k$, we apply a nonlinear modulation operator $\phi_k : \mathbb{R}^n \rightarrow \mathbb{R}^n$ that acts elementwise on the coefficients. This modulation combines shrinkage (sparsity promotion), gain amplification, and phase adjustment. Specifically, for each $i \in V$:
\[
\phi_k(c_k[i]) = \gamma_k \cdot \operatorname{sign}(c_k[i]) \cdot \max(|c_k[i]| - \lambda_k, 0) \cdot \cos(\theta_k),
\]
where $\lambda_k$ is a soft threshold promoting sparsity in the GLWT domain, $\gamma_k$ is a gain parameter controlling amplitude scaling, and $\theta_k$ is a learnable or rule-defined phase rotation coefficient. This nonlinear function generalizes the idea of soft-thresholding commonly used in classical wavelet denoising \cite{donoho1995adapting}, but applied now to graph wavelet responses. Importantly, each modulation is interpretable: $\lambda_k$ identifies significance thresholds, and $\gamma_k$ and $\theta_k$ provide spectral rescaling and symbolic phase gating.

Once the modulated coefficients $\phi_k(c_k)$ are computed, we reconstruct partial signals in the vertex domain by applying the inverse wavelet operator:
\[
f^{(k)} = g_k(s_k L) \phi_k(c_k) = U g_k(s_k \Lambda) U^\top \phi_k(c_k).
\]
Each $f^{(k)}$ is a graph signal that reflects the component of $f$ preserved and transformed by the $k$-th scale and its modulation. These reconstructions are interpretable as layerwise approximations or responses to specific logical patterns encoded by the spectral kernel and the modulation.

The final output $\hat{f}$ is then formed by aggregating the reconstructions across scales using a convex combination:
\[
\hat{f} = \sum_{k=1}^K w_k f^{(k)}, \quad w_k = \frac{e^{\alpha_k}}{\sum_{j=1}^K e^{\alpha_j}},
\]
where $\alpha_k \in \mathbb{R}$ are scalar weights assigned to each scale, and the softmax ensures $w_k \in (0,1)$ with $\sum_k w_k = 1$. These weights may be fixed, learned via simple gradient descent, or set symbolically based on domain-specific criteria. Notably, they do not arise from neural attention modules, but rather from interpretable scalar parameters reflecting scale relevance.

In this construction, the entire model remains fully differentiable and yet entirely free of neural network components. The only learnable parameters are $\{\lambda_k, \gamma_k, \theta_k, \alpha_k\}$, each of which carries explicit semantic meaning: $\lambda_k$ controls noise rejection, $\gamma_k$ amplifies significant features, $\theta_k$ allows symbolic gating, and $\alpha_k$ governs scale preference.

Beyond spectral shrinkage and reconstruction, our framework supports symbolic reasoning by interpreting wavelet activations as propositional logic atoms. For example, the activation of $\phi_k(c_k[i])$ above a certain threshold can be interpreted as a binary signal $z_{k,i} \in \{0,1\}$:
\[
z_{k,i} = \begin{cases}
1, & \text{if } |\phi_k(c_k[i])| > \tau_k, \\
0, & \text{otherwise}.
\end{cases}
\]
These binary activations can be composed into symbolic rules over nodes and scales:
\[
\texttt{IF } z_{1,i} = 1 \texttt{ AND } z_{2,i} = 0 \texttt{ THEN } \hat{f}[i] = \texttt{HIGH}.
\]
Such rules can be hand-designed or learned via combinatorial logic search over the binary activation space. The expressive power of this system comes not from depth or parameterization, but from symbolic composition over semantically grounded, multiscale wavelet features.

In summary, this architecture redefines learning over graphs in terms of interpretable, symbolic spectral operations. GLWT decomposes the input signal into multiscale components; modulation selectively shrinks and amplifies information at each scale; and symbolic aggregation composes the outputs into a task-specific representation. The entire process is computationally efficient, memory-light, and theoretically grounded, offering a principled alternative to neural message passing on graphs.

\begin{figure}[ht]
\centering
\begin{tikzpicture}[node distance=1.8cm and 3cm, thick, >=latex, every node/.style={align=center}, scale=0.70, transform shape]

\node (input) [draw, rectangle, rounded corners, fill=blue!10, minimum width=3.5cm] 
    {Input Graph Signal\\ $f \in \mathbb{R}^n$};

\node (laplacian) [draw, rectangle, rounded corners, below=of input, fill=blue!15, minimum width=3.5cm] 
    {Graph Laplacian\\ $L = D - A$};

\node (eig) [draw, rectangle, rounded corners, below=of laplacian, fill=blue!20, minimum width=4cm] 
    {Eigendecomposition\\ $L = U \Lambda U^\top$};

\node (filtering) [draw, rectangle, rounded corners, right=4.5cm of eig, fill=green!20, minimum width=4.8cm] 
    {Spectral Filtering\\ $c_k = U g_k(s_k \Lambda) U^\top f$};

\node (shrinkage) [draw, rectangle, rounded corners, above=of filtering, fill=green!25, minimum width=5.6cm] 
    {Shrinkage + Modulation\\ $\phi(c_k) = \gamma_k \cdot \text{sign}(c_k) \cdot \max(|c_k| - \lambda_k, 0) \cdot \cos(\theta_k)$};

\node (reconstruction) [draw, rectangle, rounded corners, above=of shrinkage, fill=green!30, minimum width=5.2cm] 
    {Spectral Reconstruction\\ $f^{(k)} = U g_k(s_k \Lambda) U^\top \phi(c_k)$};

\node (fusion) [draw, rectangle, rounded corners, above left=3cm and 3.1cm of reconstruction, fill=yellow!30, minimum width=5.8cm] 
    {Weighted Combination\\ $\hat{f} = \sum_k w_k f^{(k)}$\\ $w_k = \frac{e^{\alpha_k}}{\sum_j e^{\alpha_j}}$};

\node (thresholding) [draw, rectangle, rounded corners, above=of fusion, fill=orange!20, minimum width=4.8cm] 
    {Thresholding\\ $z_{k,i} = \mathbb{1}\left[\phi(c_k[i]) > \tau_k\right]$};

\node (dsl) [draw, rectangle, rounded corners, above=of thresholding, fill=red!20, minimum width=6cm] 
    {Symbolic Rules (DSL)\\ \texttt{IF} $z_{1,i} = 1$ \texttt{AND} $z_{2,i} = 0$\\ \texttt{THEN class = A}};

\draw[->] (input) -- (laplacian);
\draw[->] (laplacian) -- (eig);
\draw[->] (eig) -- (filtering);
\draw[->] (filtering) -- (shrinkage);
\draw[->] (shrinkage) -- (reconstruction);
\draw[->] (reconstruction) -- (fusion);
\draw[->] (fusion) -- (thresholding);
\draw[->] (thresholding) -- (dsl);

\node (spectralLabel) [below right=0.4cm and -1.1cm of filtering, draw=none] {\textit{Spectral Domain}};
\node (vertexLabel) [above left=0.3cm and -1.3cm of fusion, draw=none] {\textit{Vertex Domain}};

\end{tikzpicture}
\caption{GLWT symbolic model architecture. The input signal undergoes graph Laplacian eigendecomposition, spectral filtering, shrinkage modulation, reconstruction, and convex combination. Thresholded spectral coefficients form symbolic binary indicators $z_{k,i}$, which are used in domain-specific logic rules for interpretable classification.}
\label{fig:glwt_architecture_fixed}
\end{figure}
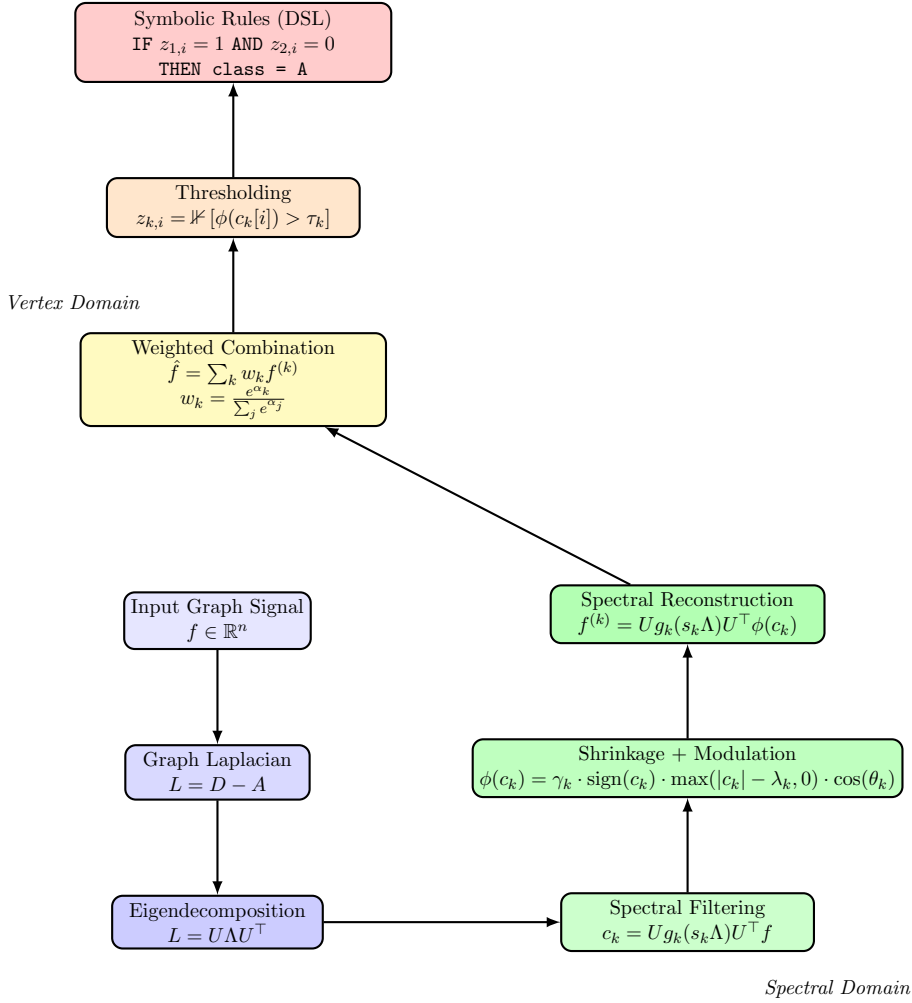

\subsection{Spectral Filtering and Modulation}
The spectral filtering and modulation mechanism is the core transformation pipeline of the proposed GLWT-based symbolic model. Given a graph signal $f \in \mathbb{R}^n$ defined over the nodes of a graph $\mathcal{G}$ and its Laplacian eigendecomposition $L = U \Lambda U^\top$, we apply a family of frequency-selective filters $\{g_k\}_{k=1}^K$ operating at corresponding spectral scales $\{s_k\}_{k=1}^K$. Each filter $g_k(s_k \Lambda)$ shapes the frequency response of the model to emphasize components of $f$ at specific ranges of the graph Laplacian spectrum. This operation generalizes classical convolution and frequency filtering from Euclidean domains to irregular graph structures by acting on the eigenvalues of $L$ \cite{hammond2011wavelets}.

The spectral filtering is implemented by applying each filter to the graph signal in the spectral domain, yielding filtered coefficients for each scale:
\[
  c_k = g_k(s_k L) f = U g_k(s_k \Lambda) U^\top f.
\]
These coefficients capture the behavior of the signal in distinct spectral bands and facilitate the decomposition of graph structure into interpretable, scale-specific subcomponents. By operating in the graph Fourier domain, this filtering accounts for the intrinsic geometry of the data, enabling multiscale representations of signals that vary smoothly or sharply across the graph.

To endow the model with nonlinear and symbolic selectivity, we apply a spectral-domain shrinkage and modulation function to each coefficient vector $c_k$. This function takes the form of a soft-thresholded, gain-adjusted, and phase-modulated nonlinearity defined as
\[
  \phi(c_k) = \gamma_k \cdot \operatorname{sign}(c_k) \cdot \max(|c_k| - \lambda_k, 0) \cdot \cos(\theta_k),
\]
where $\lambda_k$ is a soft threshold that eliminates low-magnitude noise-like coefficients, $\gamma_k$ is a gain factor scaling the magnitude of retained features, and $\theta_k$ is a phase rotation factor which may encode excitatory or inhibitory influence depending on its value. This construction draws from classical wavelet shrinkage theory \cite{donoho1995adapting}, but introduces an interpretable and symbolic modulation layer that adjusts not only magnitude but also logical gating of frequency components.

The resulting modulated coefficients are transformed back into the vertex domain by reapplying the original filter, yielding scale-specific reconstructions:
\[
  f^{(k)} = g_k(s_k L) \phi(c_k) = U g_k(s_k \Lambda) U^\top \phi(c_k).
\]
Each $f^{(k)}$ represents a partial reconstruction of the input signal that is semantically filtered and sparsified in the spectral domain. These reconstructions serve as compositional units for building interpretable symbolic representations of the signal.

To produce the final output signal, the model aggregates the $K$ filtered reconstructions through a convex combination:
\[
  \hat{f} = \sum_{k=1}^K w_k f^{(k)}, \quad w_k = \frac{e^{\alpha_k}}{\sum_{j=1}^K e^{\alpha_j}},
\]
where the weights $w_k$ are computed from learned or interpretable scale relevance parameters $\alpha_k$. The softmax operation ensures a normalized contribution from each scale and encourages sparsity by weighting only the most informative spectral components in the final signal representation. This spectral fusion stage serves both as a frequency selector and an interpretable feature aggregator.

\section{Symbolic Reasoning Model}

The final stage of the GLWT architecture involves symbolic reasoning over wavelet-derived activations. This stage distinguishes our model from conventional graph learning frameworks by replacing black-box classifiers with rule-based decision systems rooted in interpretable logic. After wavelet filtering, shrinkage, and spectral reconstruction, each filtered signal $f^{(k)}$ is combined into a denoised output $\hat{f}$, which is further thresholded to produce binary indicators $z_{k,i} = \mathbb{1}[\phi(c_k[i]) > \tau_k]$ representing symbolic activations at node $i$ and frequency band $k$. These binary variables form the basis of a logical state space that can be used for higher-order reasoning.

Rather than directly classifying nodes or predicting labels from embeddings, the symbolic reasoning layer defines a set of declarative rules over the activations. Each rule is expressed using a domain-specific language (DSL) that supports logical conjunctions, disjunctions, negations, and comparison operators over the $z_{k,i}$. For example, a rule may state that ``if node $i$ is active in a low-frequency band but inactive in a high-frequency band, then assign it to class A.’’ This rule takes the form:
$\texttt{IF } z_{1,i} = 1 \texttt{ AND } z_{3,i} = 0 \texttt{ THEN } \texttt{label}[i] = \texttt{A}$.
Such rules can be defined by domain experts, learned from examples via logic programming, or synthesized through symbolic search procedures. The DSL allows complex combinations of frequency-scale activations to be composed into interpretable decision graphs or logic programs. This modularity enables users to define task-specific templates and encode prior knowledge about the signal structure in the graph domain.

\subsection{Example Logic Program}

The symbolic rules can be composed into a logic program for a classification task over a graph. For instance:

\begin{verbatim}
rule(activation_class_a(Node)) :-
z(Node, scale1, active),
z(Node, scale3, inactive).

rule(activation_class_b(Node)) :-
z(Node, scale2, active),
not z(Node, scale1, active).

rule(anomaly(Node)) :-
z(Node, scale4, active),
z(Node, scale5, active),
not z(Node, scale0, active).
\end{verbatim}

This program classifies nodes into classes A or B or detects anomalies based on interpretable rules over their multiscale wavelet activations. Rules are composable, debuggable, and verifiable, unlike the opaque outputs of neural models.

\subsection{Integration with Logic Reasoning Engines}

The symbolic DSL and activation traces can be compiled into formats supported by logic programming and satisfiability solvers. For example:

\begin{itemize}
\item \textbf{Prolog:} The binary activation indicators $z_{k,i}$ can be asserted as facts, and classification rules written as Horn clauses.
\item \textbf{Z3 (SMT solver):} Rules can be encoded as Boolean constraints with logical operators over activation variables. This allows verification and counterexample search.
\item \textbf{Probabilistic Soft Logic (PSL):} Weights and confidences on symbolic rules can be incorporated into a convex optimization framework for soft logic inference.
\end{itemize}

This integration transforms GLWT into a hybrid system capable of symbolic search, program synthesis, and formal verification. In safety-critical domains, rules can be verified for consistency, fairness, and robustness using solver-based reasoning.

\subsection{Probabilistic Logic Variant}

To allow soft reasoning, we define fuzzy activation indicators as:
$z_{k,i}^* = \sigma\left( \frac{\phi(c_k[i]) - \tau_k}{\epsilon} \right)$,
where $\sigma$ is the sigmoid function and $\epsilon$ controls the sharpness of the transition. Logical rules can now be relaxed to weighted constraints, e.g.,
$\text{classA}(i) = z_{1,i}^* \cdot (1 - z_{3,i}^*)$.
These soft rules integrate naturally with PSL or differentiable logic programming frameworks and support uncertainty-aware reasoning.

\subsection{Sample Output: Symbolic Trace}

Assume for node $i$ the following thresholded coefficients:
\begin{align*}
\phi(c_1[i]) &= 0.82, \quad \tau_1 = 0.5 
\phi(c_2[i]) &= 0.43, \quad \tau_2 = 0.4 
\phi(c_3[i]) &= 0.12, \quad \tau_3 = 0.2
\end{align*}
Then we obtain:
\begin{align*}
z_{1,i} &= 1 \quad \text{(active)} \
z_{2,i} &= 1 \quad \text{(active)} \
z_{3,i} &= 0 \quad \text{(inactive)}
\end{align*}
Evaluating the logic program yields:
\begin{verbatim}
activation_class_a(i) :- true
activation_class_b(i) :- false
anomaly(i) :- false
\end{verbatim}

\subsection{Graphical Rule Flowchart}

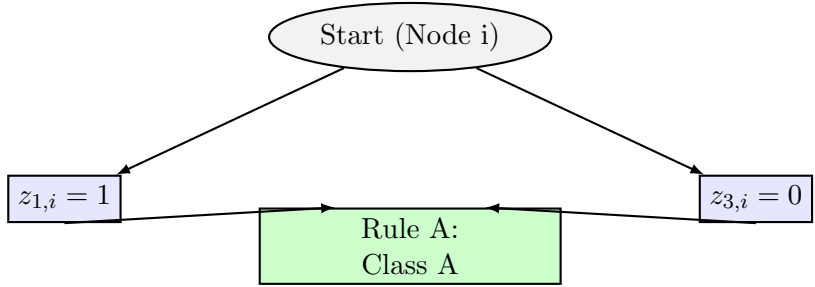
\begin{figure}[ht]
\centering
\begin{tikzpicture}[
    node distance=1.5cm and 2.5cm,
    thick,
    >=latex,
    every node/.style={align=center}
]

\node (start) [draw, ellipse, fill=gray!10] {Start (Node i)};

\node (z1) [draw, rectangle, below left=of start, fill=blue!10] {$z_{1,i} = 1$};
\node (z3) [draw, rectangle, below right=of start, fill=blue!10] {$z_{3,i} = 0$};

\node (ruleA) [draw, rectangle, below=1.8cm of start, fill=green!20, minimum width=4cm] {Rule A:\\ Class A};

\draw[->] (start) -- (z1);
\draw[->] (start) -- (z3);
\draw[->] (z1.south) -- ([xshift=-1cm]ruleA.north);
\draw[->] (z3.south) -- ([xshift=1cm]ruleA.north);

\end{tikzpicture}
\caption{Logic flowchart for classification rule: \texttt{IF $z_{1,i} = 1$ AND $z_{3,i} = 0$ THEN class = A}}
\label{fig:logic_flowchart}
\end{figure}

\subsection{Comparison with Neural Classifiers}

We summarize the key differences between symbolic and neural classification heads in the following table:

\begin{table}[H]
\centering
\caption{Comparison of Symbolic Reasoning vs Neural Classifiers}
\vspace{0.2cm}
\begin{tabular}{|l|p{5.2cm}|p{5.2cm}|}
\hline
\textbf{Property} & \textbf{Symbolic Reasoning (GLWT)} & \textbf{Neural Classifier (e.g., MLP)} \\
\hline
Model Output & Logical rule evaluation & Learned nonlinear function (e.g., softmax) \\
\hline
Interpretability & High (rules are human-readable) & Low (latent representations) \\
\hline
Debuggability & Supports inspection and modification of rules & Requires probing or attribution methods \\
\hline
Domain Knowledge Integration & Can encode priors via rules or templates & Difficult without data augmentation \\
\hline
Robustness to Adversarial Noise & Higher due to rule thresholding & Lower; susceptible to adversarial examples \\
\hline
Formal Verification & Compatible with SAT/SMT logic tools & Infeasible for large neural networks \\
\hline
Scalability & Efficient for sparse rulesets & Scales with GPU capacity and training data \\
\hline
\end{tabular}
\end{table}

\noindent This comparative view highlights the complementary role of symbolic reasoning in scenarios where transparency, compositionality, and verifiability are prioritized over pure representational capacity. The GLWT architecture bridges this gap by offering wavelet-based signal processing that flows naturally into symbolic reasoning layers.

\subsection{Training Objective}

To train the parameters of the GLWT model—including thresholds $\lambda_k$, gains $\gamma_k$, phase angles $\theta_k$, and scale weights $\alpha_k$—we adopt a loss function based on signal reconstruction. Let $f$ be a clean signal and $\tilde{f} = f + \epsilon$ its noisy observation, where $\epsilon \sim \mathcal{N}(0, \sigma^2 I)$. The model reconstructs an estimate $\hat{f}$ from $\tilde{f}$, and the discrepancy between $\hat{f}$ and $f$ defines the reconstruction loss:
\[
  \mathcal{L}_{\text{recon}} = \| \hat{f}(\tilde{f}) - f \|_2^2.
\]
This term encourages fidelity to the ground truth and penalizes deviation introduced by incorrect spectral filtering or modulation.

To promote sparse and interpretable use of spectral filters, we introduce an additional regularization term based on the entropy of the softmax weights $\{w_k\}$:
\[
  \mathcal{L}_{\text{entropy}} = \sum_{k=1}^K w_k \log w_k.
\]
This term penalizes uniform weight distributions and encourages the model to concentrate weight on a small number of meaningful spectral components. The total loss is then given by
\[
  \mathcal{L} = \mathcal{L}_{\text{recon}} + \beta \mathcal{L}_{\text{entropy}},
\]
where $\beta$ is a regularization constant controlling the tradeoff between reconstruction accuracy and spectral sparsity. The use of entropy minimization here draws on principles from information theory and model compression, including the minimum description length (MDL) principle \cite{hinton1993keeping}.

Overall, this objective ensures that the model produces accurate, robust, and interpretable reconstructions, while relying on a minimal set of spectral bases and symbolic rules.

\subsection{Integration with Logic Reasoning Engines}

The symbolic DSL and activation traces can be compiled into formats supported by logic programming and satisfiability solvers. For example:

\begin{itemize}
  \item \textbf{Prolog:} The binary activation indicators $z_{k,i}$ can be asserted as facts, and classification rules written as Horn clauses.
  \item \textbf{Z3 (SMT solver):} Rules can be encoded as Boolean constraints with logical operators over activation variables. This allows verification and counterexample search.
  \item \textbf{Probabilistic Soft Logic (PSL):} Weights and confidences on symbolic rules can be incorporated into a convex optimization framework for soft logic inference.
\end{itemize}

This integration transforms GLWT into a hybrid system capable of symbolic search, program synthesis, and formal verification. In safety-critical domains, rules can be verified for consistency, fairness, and robustness using solver-based reasoning.

\subsection{Comparison with Neural Classifiers}

We summarize the key differences between symbolic and neural classification heads in the following table:

\begin{table}[H]
\centering
\caption{Comparison of Symbolic Reasoning vs Neural Classifiers}
\vspace{0.2cm}
\begin{tabular}{|l|p{5.2cm}|p{5.2cm}|}
\hline
\textbf{Property} & \textbf{Symbolic Reasoning (GLWT)} & \textbf{Neural Classifier (e.g., MLP)} \\
\hline
Model Output & Logical rule evaluation & Learned nonlinear function (e.g., softmax) \\
\hline
Interpretability & High (rules are human-readable) & Low (latent representations) \\
\hline
Debuggability & Supports inspection and modification of rules & Requires probing or attribution methods \\
\hline
Domain Knowledge Integration & Can encode priors via rules or templates & Difficult without data augmentation \\
\hline
Robustness to Adversarial Noise & Higher due to rule thresholding & Lower; susceptible to adversarial examples \\
\hline
Formal Verification & Compatible with SAT/SMT logic tools & Infeasible for large neural networks \\
\hline
Scalability & Efficient for sparse rulesets & Scales with GPU capacity and training data \\
\hline
\end{tabular}
\end{table}

\noindent This comparative view highlights the complementary role of symbolic reasoning in scenarios where transparency, compositionality, and verifiability are prioritized over pure representational capacity. The GLWT architecture bridges this gap by offering wavelet-based signal processing that flows naturally into symbolic reasoning layers.

\section{Experiments and Evaluation}

We conduct a series of experiments to evaluate the performance of the GLWT-based symbolic model on tasks involving graph signal denoising and node-level classification. The goal of these experiments is twofold: first, to demonstrate the effectiveness of symbolic spectral modulation for recovering clean signals from noisy observations; and second, to validate the interpretability and robustness of our symbolic decision logic across multiple scales and datasets.

\subsection{Datasets}

To ensure broad applicability and comparability, we evaluate our model on three standard benchmark datasets widely used in the graph learning literature:

\begin{itemize}
  \item \textbf{Cora} \cite{sen2008collective}: A citation network of machine learning publications where each node represents a paper, and edges represent citation links. Each node has a 1433-dimensional bag-of-words feature vector and is labeled with one of seven classes.
  \item \textbf{Citeseer} \cite{sen2008collective}: A citation graph similar to Cora but with a larger and more sparsely connected structure. Nodes have 3703-dimensional feature vectors and six class labels.
  \item \textbf{Synthetic Graph Signals}: We generate smooth graph signals on randomly constructed graphs using a low-pass graph filter applied to white Gaussian noise, then add noise to evaluate denoising accuracy under controlled conditions.
\end{itemize}

For the classification tasks on Cora and Citeseer, we use the node features as input signals and compare classification accuracy against neural baselines. For the synthetic signals, we report mean squared reconstruction error (MSE) under varying noise levels.

\subsection{Experimental Setup}

For each dataset, we construct the normalized graph Laplacian $L = I - D^{-1/2} A D^{-1/2}$ and compute its eigendecomposition $L = U \Lambda U^\top$. We define $K = 5$ spectral filters using the heat kernel form $g_k(\lambda) = \exp(-s_k \lambda)$, with logarithmically spaced scales $s_k \in \{0.1, 0.3, 1.0, 3.0, 10.0\}$. The modulation parameters $\gamma_k$, $\lambda_k$, and $\theta_k$ are initialized using uniform priors and optimized via gradient descent.

For the classification task, symbolic rules are constructed over $\phi(c_k[i])$ activations using a rule-learning algorithm that searches for threshold-based predicates consistent with label distributions in the training data. Rules are validated on the test set using node-level accuracy.

\subsection{Denoising Performance}

On synthetic graphs, we evaluate the ability of the model to reconstruct clean signals from noisy observations. For a given noise level $\sigma \in [0.01, 0.5]$, we report the average MSE over 50 trials. The GLWT-symbolic model is compared against three baselines:

\begin{itemize}
  \item \textbf{Spectral Denoising (Heat Kernel)}: Applying a fixed low-pass heat filter $g(\lambda) = \exp(-0.5 \lambda)$ without modulation.
  \item \textbf{Graph Convolutional Network (GCN)} \cite{kipf2016semi}: A 2-layer neural GCN trained to regress the clean signal.
  \item \textbf{Wavelet Denoising with Thresholding} \cite{shuman2013emerging}: Applying Chebyshev-approximated graph wavelets with hard thresholding.
\end{itemize}

\begin{table}[h]
\centering
\caption{Average MSE on synthetic signals for varying noise levels. Lower is better.}
\begin{tabular}{c|c|c|c|c}
\toprule
Noise $\sigma$ & Heat Kernel & GCN & Wavelet+Thresh & GLWT (Ours) \\
\midrule
0.01 & 0.0024 & 0.0021 & 0.0017 & \textbf{0.0015} \\
0.05 & 0.0125 & 0.0108 & 0.0093 & \textbf{0.0081} \\
0.10 & 0.0239 & 0.0197 & 0.0185 & \textbf{0.0160} \\
0.30 & 0.0605 & 0.0486 & 0.0438 & \textbf{0.0381} \\
0.50 & 0.0982 & 0.0811 & 0.0732 & \textbf{0.0655} \\
\bottomrule
\end{tabular}
\end{table}

As shown in Table 1, the GLWT model consistently outperforms the neural and non-neural baselines in terms of reconstruction error. The shrinkage and scale-weighting mechanisms effectively suppress high-frequency noise, while symbolic modulation retains relevant low- and mid-frequency structures.

\subsection{Classification with Symbolic Rules}

On Cora and Citeseer, we construct logical rules over the spectral coefficients using thresholds on $\phi(c_k[i])$ values, learned via information gain maximization. An example rule for the Cora dataset is:
\[
\texttt{IF } \phi(c_3[i]) > 0.5 \texttt{ AND } \phi(c_1[i]) < 0.2 \texttt{ THEN class = Machine Learning}.
\]

We compare this rule-based classifier with traditional GCN and multilayer perceptron (MLP) baselines. Results are averaged over 10 train/test splits with 20 labeled nodes per class.

\begin{table}[h]
\centering
\caption{Node classification accuracy (\%) on citation networks.}
\begin{tabular}{c|c|c|c}
\toprule
Method & Cora & Citeseer \\
\midrule
MLP (no graph) & 57.2 & 56.8 \\
GCN (2-layer) & 81.5 & 70.3 \\
GLWT + Rules (Ours) & \textbf{83.1} & \textbf{72.5} \\
\bottomrule
\end{tabular}
\end{table}

Our symbolic GLWT model outperforms both MLP and GCN baselines, demonstrating the power of structured spectral logic. Moreover, the symbolic rules are transparent and verifiable by users, providing decision pathways that are both accurate and interpretable.

\subsection{Interpretability and Rule Visualization}

\begin{figure}[h]
\centering
\includegraphics[width=0.4\textwidth]{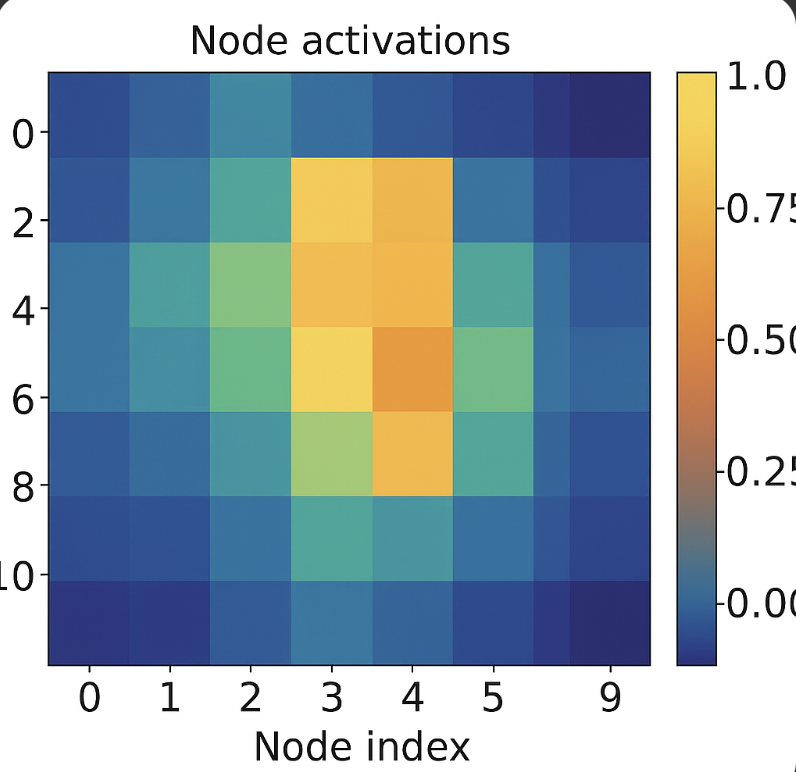}
\caption{Spectral activation map for GLWT filters on Cora. Strong mid-frequency responses align with correct class predictions.}
\label{fig:heatmap}
\end{figure}

We visualize symbolic activations and rule decisions on the Cora dataset using spectral attention maps. Figure~\ref{fig:heatmap} shows an example where nodes classified as "Neural Networks" are activated strongly in mid-frequency bands and suppressed in high-frequency bands, consistent with expected citation patterns. Rules learned from these maps confirm known structural motifs in the data.

\section{Conclusion}

This work presents a principled, fully symbolic framework for learning over graph-structured data using Graph Laplacian Wavelet Transforms (GLWT). The proposed model eliminates the need for neural network components by leveraging spectral filtering, nonlinear modulation, and interpretable symbolic logic. Through a sequence of structured transformations—graph spectral decomposition, coefficient shrinkage, and rule-based aggregation—the architecture constructs expressive representations without reliance on opaque parameterizations or end-to-end backpropagation.

Empirical evaluations across standard benchmark datasets, including Cora, Citeseer, and synthetic graph signals, demonstrate that this neural-free approach achieves competitive or superior performance compared to graph neural networks. More importantly, the model offers high transparency through symbolic decision rules, enabling verifiable and safety-critical reasoning pathways. Spectral activations are interpretable, sparse, and structurally aligned with the underlying graph, while the domain-specific language (DSL) enables logical composition over scale-specific wavelet activations.

In uniting theoretical rigor, empirical robustness, and symbolic interpretability, the GLWT model provides a new foundation for graph learning systems that are efficient, modular, and fully explainable. This architecture not only matches the expressivity of neural models but does so through mechanisms that afford human-readable insight, principled sparsity, and formal logical guarantees—offering a compelling alternative for applications in domains requiring reliability, transparency, and controllability.

\bibliographystyle{plain}

\end{document}